# Federated learning: Applications, challenges and future directions

Subrato Bharati[a,*], M. Rubaiyat Hossain Mondal[a], Prajoy Podder[a], V. B. Surya Prasath[b,c,d,e]

[a]Institute of Information and Communication Technology, Bangladesh University of Engineering and Technology, Dhaka, Bangladesh

[b]Division of Biomedical Informatics, Cincinnati Children's Hospital Medical Center, OH 45229 USA

[c]Department of Pediatrics, University of Cincinnati College of Medicine, Cincinnati, OH 45257 USA

[d]Department of Biomedical Informatics, College of Medicine, University of Cincinnati, OH 45267 USA

[e]Department of Electrical Engineering and Computer Science, University of Cincinnati, OH 45221 USA

[*]subratobharati1@gmail.com

**Abstract.** Federated learning (FL) is a system in which a central aggregator coordinates the efforts of multiple clients to solve machine learning problems. This setting allows training data to be dispersed in order to protect privacy. The purpose of this paper is to provide an overview of FL systems with a focus on healthcare. FL is evaluated here based on its frameworks, architectures, and applications. It is shown here that FL solves the preceding issues with a shared global deep learning (DL) model via a central aggregator server. This paper examines recent developments and provides a comprehensive list of unresolved issues, inspired by the rapid growth of FL research. In the context of FL, several privacy methods are described, including secure multiparty computation, homomorphic encryption, differential privacy, and stochastic gradient descent. Furthermore, a review of various FL classes, such as horizontal and vertical FL and federated transfer learning, is provided. FL has applications in wireless communication, service recommendation, intelligent medical diagnosis systems, and healthcare, all of which are discussed in this paper. We also present a thorough review of existing FL challenges, such as privacy protection, communication cost, system heterogeneity, and unreliable model upload, followed by future research directions.

**Keywords:** Artificial intelligence, Federated learning, Computing methodologies, Machine learning, Privacy protection, Healthcare

# 1  Introduction

With the advent of big data, we are no longer preoccupied with the sheer volume of data. Privacy and security of data is the most pressing issue that has to be addressed. The public is increasingly concerned about data security, and the leaking of data is never a minor issue [1-4]. Individuals, groups, and society as a whole are all working to improve privacy and data security. GDPR [5], the EU's new data protection regulations, went into effect on May $25^{th}$, 2018, and is aimed at safeguarding the privacy and security of EU citizens' personal data. Because of this, operators must explain their user agreements properly and cannot mislead or convince consumers to relinquish their privacy rights. Users must provide their consent before operators may train models. It also enables users to remove their personal information. Network operators are prohibited from releasing, tampering with, or destroying personal information they gather under China's CyberSecurity Law [6] and the General Principles of Civil Law of the People's Republic of China [7]. It is confirmed that the mentioned contract properly describes the extent of the information to be exchanged as well as the requirements for the protection of data. To varying degrees, the rules and regulations create novel difficulties for processing the classic data paradigm of artificial intelligence (AI) [150]. Because AI relies heavily on data, model training is impossible without it. As a result, there are many data islands. Data islands may be solved by centralizing the processing of the data. Centralized data collection, uniform processing, cleansing, and modeling are common approaches to data processing.

When data is collected and processed, it is almost always exposed to the outside world. Regulations have improved, but it is becoming more difficult to obtain data to train algorithms as a result. In the age of artificial intelligence, numerous individuals are arguing how to solve the problem of data islands lawfully. Traditional data analytics methodologies are already overburdened by many rules when it comes to resolving the problem of data silos. The challenge of data islands shifts the emphasis of research in federated learning. As with previous methods, the training data must be focused on one server when using the centralized technique to create a machine learning (ML) model. Privacy of data requirements have made it more difficult to use the central training approach, which might potentially expose private information and intrude on the privacy of the data owner. If smartphone users wish to train ML algorithms using their own information in centralized training settings, it is evident that the quantity of data they have is not adequate. In order to implement federated learning, users must first transfer between their personal smartphone data and the hostess server so that the ML models may be trained using the data that has been collected. Individual users from disparate geographical locations may collaborate to develop ML models applying federated learning (FL), a distributed training approach that differs from centralized training. The device may be used to store any kind of personal data, including very sensitive personal information. Users may benefit from FL without disclosing between sensitive personal information or data and a hostess server [8-11]. For artificial intelligence, FL opens novel research avenues. Federated learning offers a new way to train customized models without jeopardizing the privacy of the users. Artificial intelligence chipsets have increased the computational capabilities of client devices. The central server is rapidly being replaced by terminal equipment for artificial intelligence model training. In order to secure private information, federated learning uses the terminal device's processing capabilities to train the algorithm, which effectively prevents private data from being exposed at the time of data transfer. Because there are so many mobile devices and other devices in many domains, federated learning can take full advantage of the enormous quantity of relevant dataset resources that are available. To secure user privacy, federated learning uses techniques i.e., *k*-order anonymity and differential privacy, although it differs significantly from these more conventional methods. Because attackers can not get their hands on the source data, federated learning primarily safeguards the privacy of its users. All of these measures ensure that federated learning does not compromise user privacy or violate GDPR or other laws at the data level. Based on the distribution of data, it is possible to classify federated learning into horizontal, vertical, and transfer learning. Data sets that share a large number of user attributes but differ in their number of users may be combined using horizontal federated learning (HFL). Even when the user characteristics on the two datasets do not match up very well, vertical federated learning may still be used. Use transfer learning if the two datasets do not have enough data or tags to train a model on. Distributed ML and multiparty computing are both examples of how federated learning may be used. Distributed publication of model results is one of them and it stores data for the task of distributed computing and training purposes. It uses a trustworthy central server to disseminate data and manage resources across worker nodes so as to produce the ultimate training model more quickly. Federated learning ensures privacy by allowing users total control over their own personal information, which stresses protecting data owners' private information. When it comes to protecting the privacy of students in a federated learning environment, there are two primary options. Homomorphic encryption and secure aggregation are two prevalent techniques for encrypting data. Model parameters may also include the noise of differential privacy, which is a well-liked approach. Google's planned federated learning system [12] uses a mix of secure convergence and differential privacy to keep user information private and confidential. Homomorphic encryption protection parameters have also been used in other research [13] to protect privacy. In this work, we consider the following important aspects:

- We describe the fundamentals of the FL method as a potential solution to data privacy concerns in traditional model-centric approaches to ML. The current state of the art in FL is examined, and definitions, applications, and classifications of FL are described.
- Data partitioning, communication architecture, privacy mechanisms, systems heterogeneity, and ML models are some of the components of FL that are discussed. We figure out what is going on now and what is going to happen in the future of FL research. We also look at how FL is used in the real world and sum up its properties.
- As discussed in this article, FL is fundamentally different from learning in existing distributed environments, which require basic advances in areas like distributed optimization, large-scale ML, and privacy. To address this issue, this paper provides an overview of FL from the standpoint of data characteristics, including data privacy, data distribution, data partitioning, benchmarking. We highlight the major issues and recent advancements in the subjects of wireless communication and healthcare. Finally, we demonstrate the promise of FL approaches in healthcare and explore the future applications in FL.

## 2 Description of FL

Federated learning is a new idea that was just introduced by Google. To construct machine learning models, Google wants to use datasets that are dispersed across various devices while limiting the leakage of data. Recently, federated learning has seen advances that address statistical issues [14, 15] as well as security [12, 16]. Personalization in federated learning is also a focus of study [14, 17]. Federated learning on mobile devices involves dispersed mobile user interactions where communication costs in a large-scale distribution, uneven data distribution, and device dependability are some of the important concerns for optimization. In addition, data is partitioned horizontally in the data space by user IDs or device IDs. Data privacy in a decentralized collaborative learning scenario is an important consideration in this area of research, which is why it has been linked to in [14]. We broaden the original idea of "Federated Learning" to include all privacy-preserving decentralized collaborative machine-learning approaches in order to address collaborative learning situations within companies. An introduction to federated learning and federated transfer learning may be found in [18]. Security foundations are further examined in this article, as are their connections to other fields of study, such as multi-agent theory and privacy-preserving data mining. Federated learning is a term that encompasses data partitioning, security, and applications in this area. The federated-learning system's workflow and architecture are also discussed. Fig.1 depicts an example framework of a federated learning system.

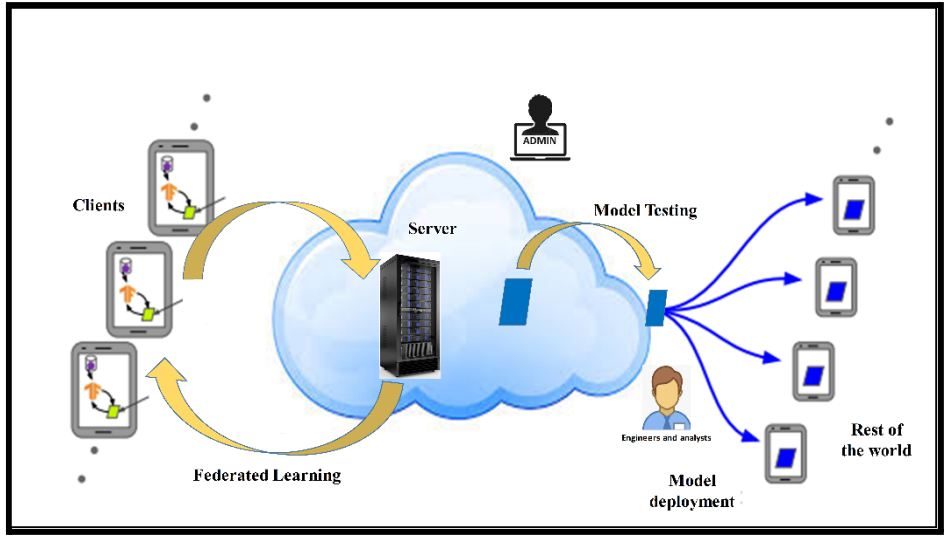

**Fig. 1.** How a FL-trained model progresses through a federated learning system's stages

## 2.1 Definition

It is important to identify the number of data owners who want to use their own data to train an AI model. Conventional practice is to combine all the data and apply $D = D_1 \cup D_2 \cup .... \cup D_N$ to train a $M_{SUM}$ model where the owners are $\{F_1, F_2, F_3, ...., F_N\}$ of $N$ data. The respective data is $\{D_1, D_2, ...., D_N\}$. In a federated learning system, the data owners work together to train a model $M_{FED}$, and no data owner $F_i$ is required to share their data $D_i$ with anyone else. It should also be mentioned that the accuracy of $V_{FED}$ is expected to be similar to that of $M_{SUM}$ (or, more precisely, $V_{SUM}$). For the sake of argument, consider the case where there is a non-negative real number ($\delta$). If

$$V_{FED} - V_{SUM} < \delta \qquad (1)$$

To put it another way, we argue that the FL algorithm suffers from an $\delta$-accuracy loss.

## 2.2. Federated Learning with its Privacy

Federated learning relies on privacy as a fundamental component. In order to give effective privacy assurances, security models and analysis are needed. In the following pages, we shortly discuss and compare several privacy methods for FL. Indirect leakage prevention methods and obstacles are also discussed.

1. SMC: Multiple parties are involved in secure multiparty computation (SMC) security models, which give security evidence in a precise framework for simulation to make sure that each party recognizes nothing but its input and output, ensuring total lack of knowledge. It is ideal to have zero knowledge, but achieving this goal frequently requires elaborate computing procedures that are not always feasible. Depending on the circumstances, it may be permissible to disclose just a portion of a person's knowledge. Reduced security needs in exchange for increased capability may be met by using SMC to construct a security model. The research [19] employed the framework for SMC to train ML algorithms using semi-honest assumptions and two servers. The author of [20] employed MPC methods for verification and model training to keep sensitive data out of the hands of users. Sharemind is one of the most advanced SMC frameworks [21]. With an honest majority [22], offered a 3PC model [23, 24] that examined security under both honest and malevolent assumptions. In order to complete these projects, non-colluding servers must discreetly exchange participant data.

2. Homomorphic Encryption: To protect user data privacy during machine learning, homomorphic encryption [25] is also used, as shown in [26-28]. There is no transmission of information or guessing of data by the other side, unlike differential privacy protection. To put it another way, there is minimal risk of raw data loss because of the encryption. The centralization and training of data in the cloud have recently been done using homomorphic encryption [29, 30]. Additive homomorphic encryption [31] is frequently employed, and polynomial approximations must be performed in machine learning algorithms to assess non-linear functions [32, 33].

3. Differential Privacy: For data privacy protection, another line of study employs the methods of *k*-anonymity [34] or differential privacy. For example, *K*-anonymity, diversification, and differential privacy all employ noise or generalization approaches to conceal some critical qualities until a 3rd party cannot differentiate the person, thus making the data hard to recover for user privacy purposes. A trade-off between privacy and accuracy often exists in these procedures, which is why the data must still be sent elsewhere. For federated learning, the authors of [16] came up with a way to keep client-side information safe by hiding the client's involvement in the training process.

4. Optimization Approach: Stochastic gradient descent (SGD) is an optimization approach that uses stochastic gradient descent (SGD) to update parameters. However, there is no assurance of security, and the gradients' leakage can really leak vital information in data [35] when revealed to data structure, for example, in the case of picture pixels. A number of researchers have taken this into consideration when a member of a federated learning system deliberately assaults another member by giving them a backdoor into the system to learn the data of others. A novel "constrain-and-scale" model-poisoning approach is proposed to limit data poisoning in [36]. The authors show that concealed back doors may be inserted into a combined global model. Researchers in [37] found that the training data utilized by multiple parties in collaborative learning may be exposed to inference attacks because of possible flaws in collaborative machine learning systems. Members and attributes linked to a portion of the training data may be inferred by an adversarial participant. Other alternative defenses were also suggested. As the authors explain in [38], gradient exchanges between multiple parties might represent a security risk, as the authors suggest a safe form of the gradient descent approach. Tolerance for Byzantine employees, they demonstrate, is up to a certain percentage constant. It has been suggested that a blockchain platform may be used to facilitate federated learning. Using a blockchain to share and verify the updates of local-learning models, researchers in [39] proposed a blockchain federated-learning (BlockFL) architecture. They have taken into account network scalability, resilience, and optimum block creation.

**2.3. Classification of Federated Learning**

According to the distribution features of the data, federated learning may be categorized. Assume that the data matrix $D_i$ represents the information owned by each individual data owner, i.e., each sample and each characteristic are represented by a row and a column, respectively, in the matrix. At the same time, label data may be included in certain datasets as well. For example, we call the sample ID space *I*, the feature space *X* and the label space *Y*. When it comes to finance, labels may represent the credit of customers; when it comes to marketing, labels can represent the desire of customers to buy; and when it comes to education, labels can represent students' degrees. The training dataset includes the features *X*, *Y*, and IDs *I*. Federated learning may be classified as horizontally, vertically, or as federated transfer learning (FTL) depending on how the data is dispersed among the many parties in the feature and sample ID space. We cannot guarantee that the sample ID and feature spaces of the data parties are similar. Several FL frameworks for a 2-party situation are shown schematically in Fig. 2.

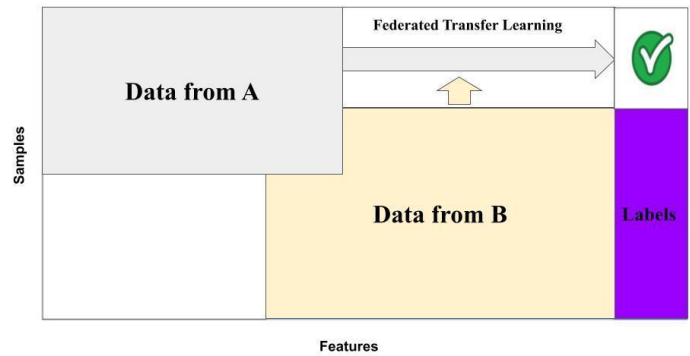

(a)

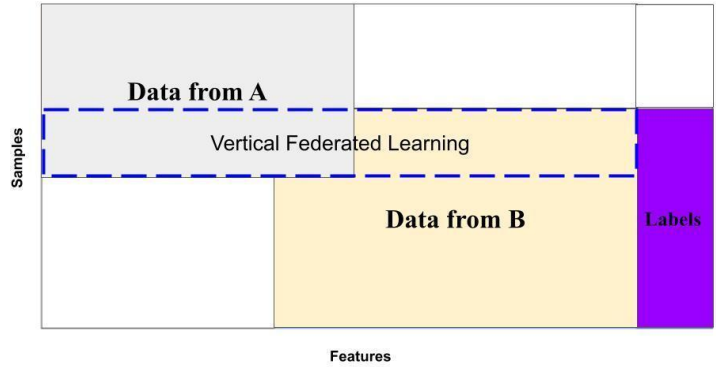

(b)

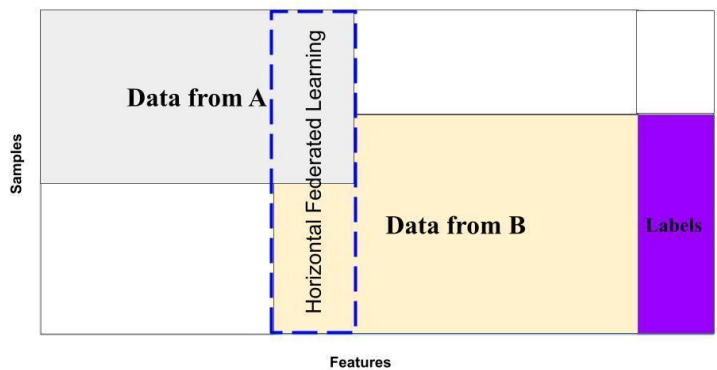

(c)

**Fig. 2.** Classification of FL where (a) Federated Transfer Learning, (b) Vertical Federated Learning, and (c) Horizontal Federated Learning

### 2.3.1 Federated Transfer Learning (FTL)

Federated transfer learning (in Figure 2(a)) is suitable while two datasets differ not only just in sample size but also in feature space. Consider a bank in China and an e-commerce firm in the United States as two separate entities. The small overlap between the user populations of the two institutions is due to geographical constraints. However, only a tiny fraction of the feature space from both companies overlaps as a result of the distinct enterprises. For example, transfer-learning [40] may be used to generate solutions of problems for the full dataset and features under a federation. Specifically, a typical portrayal across the 2 feature spaces is learnt by applying restricted general sample sets as well as then used to produce prediction results for samples with just one-sided features. There are difficulties that FTL addresses that cannot be addressed by current federated learning methods, which is why it is an essential addition to the field.

$$X_i \neq X_j, Y_i \neq Y_j, I_i \neq I_j \forall D_i, D_j, i \neq j \qquad (2)$$

**Security Definition for FTL:** Two parties are normally involved in a federated transfer learning system. Due to its protocols' being comparable to vertical federated learning, the security definition for vertical federated learning may be extended here, as will be illustrated in the next.

## 2.3.2 Vertical Federated Learning (VFL)

Machine-learning techniques for vertically partitioned data have been suggested that preserve privacy, including gradient descent [41], classification [42], secure linear regression [43-45], association rule mining [46], and cooperative statistical analysis [47]. Authors of [48, 49] have presented a VFL method for training a logistic regression model that preserves individual privacy. Entity resolution and learning performance were investigated by the authors, who used Taylor approximation to approximate gradient and loss functions in order to provide homomorphic encryption for privacy-preserving computations. While 2 datasets share the same space of sample ID but vary in feature space, VFL, also known as feature-based FL (Figure 2(b)), may be used. An ecommerce firm and a bank are two examples of businesses in the same city that operate in quite different ways. The intersection of their user spaces is enormous since their user sets are likely to include most of the inhabitants in the region. Banks and e-commerce, on the other hand, keep track of their customers' income and spending habits and credit ratings; therefore, their feature sets are vastly different. Assume that both parties seek a product purchase prediction model based on product and user data. These distinct characteristics are aggregated, and the training loss and gradients are computed to develop a model that incorporates data from both parties jointly. In a federated learning system, every participating party has the same identity and position, and the federal method helps everybody build a "common wealth" plan.

**Security Definition of VFL:** Participant honesty and curiosity are assumed in a vertically federated-learning system. In a 2-party case, 2 parties are not collaborating, as well as only one is understood by an opponent. Only the corrupted client may learn data from the other client, and only that which the input or output disclose. Occasionally, a semi-honest third party (STP) is included to permit safe calculations between the parties, in which it is believed that the STP does not collude. These protocols are protected by SMC's formal privacy proof [50]. At the conclusion of the learning process, every party has just the parameters of associated models with their own unique traits left in their memory bank. As a result, the two parties must work together at the inference stage to provide a result.

### 2.3.3 Horizontal Federated Learning (HFL)

For datasets that share a feature space but vary in the number of samples, sample-based federated learning or horizontal federated learning is presented (in Figure 2(c)). There is very little overlap between the user bases of two regional banks with very divergent customer bases. Nevertheless, due to the fact that their businesses are so similar, the feature spaces are identical. Collaboratively deep learning was suggested by the authors of [51]. Participants train individually and exchange just a limited number of parameter changes. Android phone model upgrades were suggested by Google in 2017 as a horizontal federated-learning approach [52]. One user's Android phone changes the model parameters locally, then uploads them to the training the centralized model, Android cloud in concert with other owners of data. Their federated-learning approach is further supported by a safe aggregation technique that protects the aggregated user updates in privacy [12]. To protect against the central server, model parameters are aggregated using additively homomorphic encryption as described in [35]. According to [14], several sites may execute independent tasks while maintaining security and sharing knowledge by using a multitask-style FL system. High fault tolerance stragglers, and communication costs difficulties may all be addressed by their multitask learning paradigm. It was suggested in [52] to establish a safe client–server structure where data is partitioned by users as well as models developed on client devices work together to produce a global federated algorithm in the framework of an interconnected federated learning system. The model-building procedure prevents any data from leaking. The authors of [53] looked at ways to reduce transmission costs so that data from remote mobile clients may be used in the training of centrally-located models. Deep gradient compression [54] has recently been presented as a way to significantly decrease the bandwidth required for distributing large-scale training.

**Security Definition of HFL:** An honest-but-curious server is commonly assumed in a horizontally federated learning system [12, 35, 55]. That is, only the server has the ability to intrude on the privacy of the data participants it hosts. These pieces of craftsmanship serve as evidence of their own safety. Additional privacy concerns have been raised by the recent proposal of a new security model that takes malevolent users [56, 147] into account. At the conclusion of the session, all of the parameters of the universal model are made available to each and every participant.

## 3  Applications

When used in the financial, sales, and several industries where data cannot be aggregated for the training of ML models because of such factors as data security, privacy protection, as well as intellectual property rights, federated learning has a lot of promise as an innovative modeling mechanism [57, 148, 151]. Customer-specific services, such as product recommendations and sales assistance, will be provided using machine learning methods. Users' buying power, personal preferences, and product attributes are among the most important data elements in the smart retail industry. There is a good chance that these three data elements will be spread out among three distinct departments or companies in real applications. For example, a user's bank balances and personal preferences may be used to estimate their spending power, and an e-store can track the features of the things they buy. This work is dealing with two issues in this situation. As a first step, banks, social networking sites, and e-commerce sites have data walls that are tough to overcome. As a consequence, it is not possible to train a model by aggregating data. Because the data held by the 3 parties is frequently traditional, heterogeneous ML methods cannot interact with heterogeneous data directly. These issues have yet to be resolved using standard machine learning approaches, making it difficult for AI to become more widely accepted and used. The solution to these issues lies in federated and transferable learning. Federated learning allows us to construct models for all three parties without exporting company data, which not only preserves data security and privacy but also offers customers individualized, targeted services, achieving mutual advantages. In the meantime, we may use transfer learning to solve the issue of data heterogeneity as well as overcome the constraints of standard AI approaches. Because of this, we can create a big data and artificial intelligence ecosphere that is cross-company, cross-data, and cross-domain using federated learning. The federated-learning architecture may be used to query many databases without disclosing any data. Multiparty borrowing, for example, has historically been a significant source of risk in the banking sector. This is what occurs when someone stealthily takes out a loan from one bank to pay off another bank's debt. Financial stability is threatened by multiparty borrowing since a high number of illicit activities might bring down the whole financial system. Federated learning may be used to locate such people without disclosing the user lists of banks A and B to each other. Encrypting each party's user list and then taking the intersection of those encrypted lists is possible using federated learning's encryption method. Using the final result's decryption key, the decrypted list of multiparty borrowers may be obtained without revealing the identities of the "good" users. This procedure is in line with the vertical federated learning structure, as we'll see in a moment. We anticipate federated learning to have a significant impact in the field of smart healthcare as well. Although medical datasets are very confidential and sensitive, they are difficult to gather and only exist in isolated medical institutions and hospitals. This is why medical datasets are so tough to get. The present barrier in smart healthcare is due to the poor performance of ML models due to a lack of data and labels. In our view, the efficiency of ML algorithms trained on a huge medical dataset would be considerably enhanced if every medical or healthcare institution collaborated and shared their data. The most effective strategy for achieving this goal is through a combination of transfer learning (TL) and federated learning. It is possible to use TL to fill the gaps in the information or data, thereby enhancing the performance of trained models. As a result, federated transfer learning might be critical to the advancement of smart healthcare and raise human healthcare to a new level.

**3.1. Wireless communication**

Wireless networks are becoming more complicated, and it is time to move on from the old methods based on outdated models. Likewise, the growing widespread use of deep learning networks has also shifted the focus of efforts to develop new wireless network models [58, 149]. An in-depth examination of edge computing and 5G networks was conducted by Niknam et al. [59] using the essential features of FL. In order to demonstrate the reliability and security of FL in wireless communication, simulations were run on typical data sets. Using a remote parameter server and data from each device's own data collection, the authors of [60] investigated the use of FL in edge computing and wireless networks. For the optimum transmission efficiency, Tran and colleagues [61] devised a light wave power based FL model to regulate the network's physical layer through resource allocation. Even so, Ang et al. [62] provided a technique for dealing with wireless communication noise that was resilient and federated. It was reduced to the issue of parallel optimization based on the worst-case model and the anticipated model in the broadcast process and aggregation process. To solve the relevant problem of optimization, the sampling-based Service Component Architecture (SCA) algorithm and the Service-Level Agreement (SLA) algorithm can be used. The algorithm's experimental findings suggest that it has improved prediction accuracy and reduced loss. The work of [63] used FL to enhance attack detection accuracy and communication efficiency in a smart home wireless internet of things system. Savazz and his colleagues [64] suggested a server-less learning technique for applications in 5G wireless networks that exchanges model parameters via local gradient iterative computation of consistency-based mechanism and each device. In order to boost the model's communication efficiency, the authors of [65] created a hierarchical FL system for wireless heterogeneous cellular networks (HCNs).

## 3.2. Service recommendation

**1) Google keyboard**

It was announced in 2016 that Google was working on a project to build FL among the users of Android mobile [12] in order to enhance the prediction accuracy of the keyboard inputs while also protecting user privacy and security. The recommendation system's development will be aided by the language model's progress [66]. Using federated learning, it may be used for a wide range of recommendation applications. Whenever a user asks for anything, the model may deliver an immediate answer.

**2) Intelligent medical diagnosis system**

Collecting medical records from several institutions is challenging because of the need to maintain patient confidentiality. Because of this, medical data is becoming rarer. The distribution of medical resources and the diagnosis of sickness have undergone radical transformations as a result of advances in artificial intelligence. Although the processing and gathering of data presents security problems, i.e., the exposure of patients' private information [67], In light of patient requirements, the authors of [68] evaluated the current moral/legal problems and proposed ways to utilize patient data apart from revealing privacy. Medical data has two issues: a lack of data and inadequate labeling, both of which may be remedied using the current federated transfer learning. A multi-FL network based on the APOLLO network was designed and implemented by the authors of [69] using the interconnected healthcare system to collect health longitudinal real data and outcome data and to help doctors in the future diagnosis of patients by transforming evidence of medical diagnosis from real-world data.

## 3.3. Healthcare

It is not uncommon for medical FL apps to attempt classification. Classification algorithms in machine learning [70] learn how to categorize or annotate a given collection of occurrences with labels or classes. For example, the classification tasks of COVID-19 detections [71-84], cancer diagnoses [85-94] and autism spectrum disorder (ASD) [84, 95-97] are considered in a FL setting in healthcare. Other classification tasks studied include emotion recognition and human activity [98-100] and prediction of patient hospitalization [101, 102]. Table 1 provides an overview of the classification tasks used in FL research for medical applications.

Data from electronic health records (EHRs) has become an invaluable resource for biological research [103, 104], including machine learning [105]. In spite of the fact that electronic health records (EHRs) provide an enormous quantity of patient data for analysis, they have inherent biases that are distinct to each institution and so restrict the generalizability of the findings. Algorithms used to determine participation in special healthcare programs were shown to be biased against African Americans, assigning the same level of risk to healthier Caucasian patients as was observed by Obermeyer et al. [106]. Various factors, i.e., insufficient training data depiction or disparities in access to care might lead to these incorrectly calibrated algorithms. To mitigate the danger of such biased algorithms, it is apparent that the capacity to learn from EHR data that is more typical of the wide range population as well as extends beyond a particular location or hospital is essential. Regrettably, it is doubtful that data will be linked in a single repository for the purpose of research from everything at once owing to a variety of factors, such as disparate data methods and privacy issues. Even while typical data models, i.e., OMOP [107], enable more widespread replication analysis, shared data access remains a major restriction. It is possible to link medical institutions' EHR data via federated learning while maintaining the confidentiality of the data itself [108-113]. Improvements of iterative learning from varied and big medical databases may considerably increase the performance of ML models in these situations. Predictive modeling, phenotyping [53, 114], patient representation learning, and patients' similarity learning [115] have been investigated in the context of FL in healthcare. A privacy-preserving framework for patient similarity learning across institutions was described by Lee et al. [115]. To locate comparable patients in other hospitals, their approach does not require sharing any patient-level data. In a federated learning context, the authors of [114] employed models for tensor factorization to extract relevant phenotypes from vast electronic health records. In a federated approach, Liu et al. [116] studied obesity-related comorbidity phenotyping as well as patient representation learning. A technique dubbed SplitNN [117] developed by Vepakomma et al. [118] allowed for the collaborative training of DL models by health organizations without releasing model specifications or sensitive raw data. Federated learning was shown by Silva et al. [119] by comparing brain anatomical correlations between illnesses and clinical cohorts. Clinically clustering patients in significant communities that offered comparable diagnoses and geographic locations as well as concurrently training one model for each community were the methods used by Huang et al. [111].

In addition, federated learning has made it possible for doctors to get further insights into the benefits and risks of treating patients early [102, 120, 121]. The authors of [102] set out to solve the l1-regularized sparse SVM classifier in a FL environment in order to predict the hospitalized patients with heart-related disorders utilizing EHR data scattered across several data agents/sources in future. Resistance of patients' to specific treatments as well as medications, and their survival chances for certain illnesses, are being predicted by Owkin utilizing federated learning [122]. Prolonged hospital stays and in-hospital mortality have been studied by Pfohl and colleagues in the eICU Collaborative Research Database [123]. It was found that the authors of [121] explored a privacy-preserving system for the job of predicting mortality among ICU patients (ICU). In comparison to the usual centralized learning context, they found that training the model in a FL framework yielded equivalent results. Table 1 provides a summary of the research presented here.

**Table 1:** An overview of current developments in healthcare federated learning

| Ref. | ML Method | Dataset | Clients number | Problem Statement |
|---|---|---|---|---|
| [114] | TF | MIMIC-III, UCSD | 1-5 | Phenotyping |
| [119] | PCA | MIRIAD, PPMI, Biobank, UK ADNI, | 10-100 | Representation learning |
| [102] | SVM | Boston Medical Center | 5, 10 | Hospitalization prediction |
| [123] | LR, NN | Collaborative Research Dataset of eICU | 31 | Mortality prediction |
| [124] | CNN | UCI Smartphone | 5 | Activity recognition |
| [125] | NN | PhysioNet Dataset | 16, 32, 64 | Arrhythmia detection |
| [126] | LASSO, MLP, LRR | Mount Sinai COVID-19 | 5 | Mortality prediction |
| [116] | NLP | MIMIC-III | 10 | Phenotyping |
| [111] | Autoencoder | Collaborative | 5-50 | Mortality prediction |

| | | Research Dataset of eICU | | |
|---|---|---|---|---|
| [120] | RNN | Cerner Health Facts | 50 | Preterm-birth prediction |
| [121] | MLP, LR | MIMIC-III | 2 | Mortality prediction |

# 4  Challenges

FL is a new discipline, and although it has contributed in certain areas, it now faces significant hurdles in the optimization of performance; the following are the several key issues. In this section, we will explain the most important problems in tackling the distributed optimization issue. In the federated environment, these issues are unique from different classical concerns, i.e., distributed learning in typical private data analysis or data center settings.

**4.1. Privacy protection**

Privacy is a fundamental problem in FL. Instead of transmitting raw data, federated learning exchanges model gradients with the server to safeguard private information on each device. The reverse deduction of models, for example, during the process of whole training, models are leaking critical data to a third party via communication of the model. The federated network's computing load is increased as a result, even if new ways of enhancing data privacy have been developed. Private data must not be leaked during model transfer if the necessary protection of private data is to be maintained.

Federated learning systems typically raise privacy issues as a primary concern. Instead of sharing raw data, federated learning exchanges model updates, such as gradient information, to help safeguard the data created on each device. Even so, passing forward model updates to the central server or to a third party during the training process may divulge private data [127]. Secure multiparty computation (SMC) and differential privacy are emerging strategies that try to improve privacy in federated learning, although they typically do so at the expense of system efficiency or model performance [12, 127]. In order to implement private federated learning systems, it is essential to understand and balance these tradeoffs both theoretically and practically.

**4.2. Communication cost**

Communication is a major barrier in federated learning. A FL network can be made up of a large number of devices, such as a large number of mobile devices. In order to understand how to use a FL model, there can be a great deal of communication involved. Federated learning's communication costs should be taken into account as well, since network speeds cannot be guaranteed. As a result, the development of systems with great communication efficiency is required in order to create FL practicable. Communication at a high cost data created on each device must stay local due to communication being a significant bottleneck in federated networks [4]. This is also due to privacy concerns around the transmission of raw data. Many orders of magnitude slower than local processing, federated networks may include a large number of devices, such as millions of smartphones. This is due to the limited bandwidth, energy, and power available in federated networks. Therefore, it is necessary to design communication-efficient approaches that may iteratively update models and deliver short messages throughout the training process, rather than sending the complete data set across the network as part of the training. Minimize the number of rounds of communication and the size of the messages conveyed in order to further reduce communication in such a setup.

**4.3. Systems heterogeneity**

The communication and computation abilities of every device in the federated network can vary due to differences in network and hardware connectivity. Only a tiny percentage of a network's devices are active at any given time. As a result, a network with millions or billions of devices may only have a few hundred devices online at any one moment. The heterogeneity of these systems makes fault tolerance more difficult since each device may be faulty in its own right. It is also essential that networked systems can be handled resilient to offline devices and varied hardware. Due to hardware (RAM and CPU), network connection (Wi-Fi or 4G, 5G), as well as power (battery level), the abilities of every device in FL networks can vary [128]. Because of the sheer number of devices in a network and the limitations imposed by their respective systems, only a tiny percentage of those devices are usually active at any one time [4]. As a result, issues like straggler mitigation and fault tolerance are made much more difficult. Because of this, federated learning techniques must be able to handle a small number of participants, a wide range of hardware, and the occasional loss of a communication link.

**4.4. Unreliable model upload**

Nodes might intentionally or accidentally mislead the server [129] while aggregating global models in FL. Using malicious model parameters, an attacker may cause model training to go awry by altering the aggregate of the global model. Mobile devices might also upload low-quality models because of the unpredictable network environment, which would have a negative impact on FL. For FL, it is essential to avoid uploading an unstable local model.

**4.5. Statistical heterogeneity**

There are a variety of ways that devices create and gather data, such as the fact that mobile phone users often employ a different vocabulary while doing next-word predictions. There can be a fundamental statistical structure that represents the interaction between their related distributions and devices, as well as a wide variation in the quantity of data points between devices. In distributed optimization, the assumption of identically and independently distributed data production may complicate issue modeling, theoretical analysis, and empirical assessment of solutions. In this context, prominent techniques for meta learning and federated learning [130] have a strong link. For improved personalization, the heterogeneity in statistical data can be better handled using the multitask and meta-learning viewpoints, respectively.

To begin with, current FL studies only look at one of the non-IID features, such as label skew or data imbalance. But there aren't any thorough studies in the medical dataset that look at a variety of non-IID attributes all at once. Additional techniques for dealing with hybrid non-IID characteristics are expected to be discovered throughout time in the future. The hyperparameter framework search for FL is the second restriction. The process of fine-tuning hyperparameters in a machine learning model is both important and time-consuming. Hyperparameter optimization becomes much more complex when models are trained over a distributed network of diverse data silos in federated learning. Thus, the development of an automated technique or framework for selecting the appropriate FL model hyperparameters is crucial in the future.

**4.6. Federated Fairness**

Fairness researchers should take notice because federated learning presents new possibilities and problems. Federated learning, for example, may enable modeling and research using partitioned data that would otherwise be too sensitive to disclose directly by allowing datasets to be disseminated both by observation and even by characteristics [48, 117]. Fair modeling theory and practice may be advanced by increasing the variety of training data accessible to machine learning models, which is made possible by the increased availability of datasets that can be federated. Federated learning creates new issues for researchers and practitioners in terms of fairness. Contextual factors like connection quality, device kind, geographic location, use habits, and dataset size may all influence how samples are selected in a federated learning model [4]. For example, the works of [131-133] might look at how the resultant model's fairness is affected by the different sample limits and how these effects can be minimized within the federated framework in the future. Frameworks like agnostic FL [134] provide methods for preventing bias in training objectives. Fairness improvements in current federated training algorithms will be especially critical when other FL system components, such as model compression, approach their technological constraints. This will require a greater focus on improving existing algorithms' fairness. In the study of fairness, there is no one fairness criterion that is widely accepted, and numerous criteria have been shown to be incompatible. Online fairness framework and algorithms may be used to address this issue in [135]. It will be a challenge to adapt these solutions to the FL environment as well as to more improve upon them. Distributionally-robust optimization, post-shifting techniques, and constrained optimization have all been developed in the last decade to train fair classifiers in the conventional centralized machine learning scenario [136-138]. For federated learning, it's unclear whether these methodologies, which have been shown effective in centralized training, can be utilized in the context of decentralized data.

# 5    Future directions

The study of federated learning is one that is both current and continuous. This section's survey of related and current work has started to address some of the issues raised, but there are still a number of important questions that need to be answered. Our prior obstacles (cost communication, heterogeneity, statistical heterogeneity, and privacy concerns) are briefly reviewed in this part, as are new challenges such as productionizing and benchmarking in federated contexts, which are described in the next paragraphs.

**5.1. Extreme communication schemes**

The extent to which federated learning will require communication is yet to be determined. Machine learning optimization approaches may accept a lack of accuracy; in fact, this imperfection can aid generalization [139]. There has been some research on classic data center communication systems like one-shot or divide-and-conquer [140], but the behavior of these approaches in huge and highly heterogeneous networks is not well known.

**5.2. Communication reduction and the Pareto frontier**

Federated training may benefit from local updates and model compression, among other techniques. It is critical to know how these methods work together and to evaluate the accuracy vs. communication tradeoffs for each one. It is essential that the most effective methods improve at the Pareto frontier, such as obtaining better accuracy than any other method within the same preferably and, communication budget, over a broad spectrum of accuracy/communication profiles. In order to evaluate communication-reduction approaches for federated learning meaningfully, similar thorough assessments have been undertaken for efficient inference of neural networks [141].

**5.3. Novel models of asynchrony**

In distributed optimization, asynchronous approaches and bulk synchronous techniques are two of the most extensively investigated communication strategies. Worker nodes in data centers tend to be "committed" to the workload, so they are ready to "pull" their next task from the central node as soon as they "push" the results of their previous job. Although each device in a federated network is typically not committed to the activity at hand, most devices remain inactive on any given network iteration. This more realistic device-centric communication strategy should be studied because each device may determine when it "wakes up" a new job is pulled from some local computation and the central node is performed.

**5.4. Heterogeneity diagnostics**

Recently, metrics such as local dissimilarity have been used to attempt to measure statistical heterogeneity (as developed in the context of FL and utilized elsewhere in works like [142]). Prior to training, it was impossible to readily compute these metrics across the federated network. The following questions are prompted by the relevance of these metrics:

The amount of heterogeneity in federated networks is difficult to predict before they are deployed. Is there a way to measure the level of system-related heterogeneity? It is possible to use existing or new definitions of heterogeneity, both experimentally and conceptually, to develop new federated optimization algorithms with enhanced convergence.

**5.6. Granular privacy constraints**

The section titled "Privacy in Federated Learning" provides definitions for local and global privacy pertaining to all networked devices. Since it is a practical matter, it can be required to define privacy in a more fine-grained manner, as privacy limitations can vary from device to device or even inside a single device. Even though the work of [130] recently presented sample-specific (as an alternative to a user-specific design) privacy assurances, this provides a lower level of privacy in return for more accurate models, as shown by their work [130]. Future research should focus on developing strategies for dealing with mixed (sample-specific or device-specific) privacy limitations.

**5.7. Beyond supervised learning**

As a reminder, all of the approaches presented so far were designed for supervised learning, which means that they presume that each and every data in the FL network has already been labeled. Realistic federated networks may create data that is either unlabeled or poorly tagged. More sophisticated tasks, such as reinforcement learning, may include exploratory data analysis and aggregate statistics rather than just fitting a model to the available data as described in [143]. Problems beyond supervised learning in federated networks will most likely need to tackle comparable issues of scalability, heterogeneity, and privacy in order to be effectively addressed.

**5.8. Productionizing federated learning**

In addition to the primary ones described here, there are a slew of practical issues that crop up when FL is put into production. Concerns including idea drift, diurnal fluctuations [144], as well as cold-start concerns (the network expands as additional devices are added) must be addressed with caution. There are a number of actual system-related challenges in production federated learning systems, which we direct readers to in [4].

**5.9. Benchmarks**

Because federated learning is still in its infancy, we have an opportunity to define its future by ensuring that it is based on real-world circumstances, assumptions, and data sets. Finally, research groups need to improve on current implementations and benchmarking tools, such as TensorFlow Federated [145] and LEAF [146], in order to make empirical findings more reproducible and to disseminate new federated learning solutions.

# 6 Conclusions

The idea of federated learning, which provides cross-platform privacy protection, was discussed. When it comes to data privacy and security, more and more academics and businesses are coming around to the idea of federated learning as a viable paradigm. Collaborative learning may combine the integrated model and update many user models without disclosing the source data in situations when users cannot train good models due to a lack of data. However, federated learning may also offer a safe model sharing method and migrate models to particular tasks to tackle the issue of inadequate data labels when users have a limited amount of data. Compared with classical privacy-preserving learning and typical distributed data center computing, federated learning has certain distinct advantages and disadvantages. An overview of federated learning is presented here, along with examples of how it may be used in different contexts. We compiled a comprehensive overview of classic findings as well as more current studies that emphasized federated settings. We take a look at the present state of FL, which includes, but is not limited to, the healthcare sector, in this study. An increasing number of researchers are focusing on federated learning, and this article highlights recent developments and provides a wide range of open issues and concerns. Finally, we provided a list of open issues that need further investigation. A wide range of research communities will be needed to come up with answers to these issues.

## Acknowledgment

This work has been carried out at the Institute of Information and Communication Technology (IICT) of Bangladesh University of Engineering and Technology (BUET), Bangladesh. Hence, the authors would like to thank BUET for providing the support.

## References

[1] C. Zhang, X. Hu, Y. Xie, M. Gong, and B. Yu, A privacy-preserving multi-task learning framework for face detection, landmark localization, pose estimation, and gender recognition, *Frontiers in Neurorobotics,* pp. 112, 2020.
[2] M. Gong, J. Feng, and Y. Xie, Privacy-enhanced multi-party deep learning, *Neural Networks,* vol. 121, pp. 484-496, 2020.
[3] Y. Xie, H. Wang, B. Yu, and C. Zhang, Secure collaborative few-shot learning, *Knowledge-Based Systems,* vol. 203, p. 106157, 2020.
[4] K. Bonawitz *et al.*, Towards federated learning at scale: System design, *Proceedings of Machine Learning and Systems,* **1** (2019), 374-388.

[5] J. P. Albrecht, How the GDPR will change the world, *Eur. Data Prot. L. Rev.,* **2** (2016), pp. 287.

[6] M. Parasol, The impact of China's 2016 Cyber Security Law on foreign technology firms, and on China's big data and Smart City dreams, *Computer law & security review,* **34** (1) (2018), 67-98.

[7] W. Gray and H. R. Zheng, General Principles of Civil Law of the People's Republic of China, *The American Journal of Comparative Law,* **34** (4) (1986), 715-743.

[8] M. Gong, Y. Xie, K. Pan, K. Feng, and A. K. Qin, A survey on differentially private machine learning, *IEEE computational intelligence magazine,* **15** (2) (2020), 49-64.

[9] C. Zhang, Y. Xie, H. Bai, B. Yu, W. Li, and Y. Gao, A survey on federated learning, *Knowledge-Based Systems,* vol. 216, pp. 106775, 2021.

[10] T. Li, A. K. Sahu, A. Talwalkar, and V. Smith, Federated learning: Challenges, methods, and future directions, *IEEE Signal Processing Magazine,* **37**(3) (2020), 50-60.

[11] J. Xu, B. S. Glicksberg, C. Su, P. Walker, J. Bian, and F. Wang, Federated learning for healthcare informatics, *Journal of Healthcare Informatics Research,* 5(1) (2021), 1-19.

[12] K. Bonawitz *et al.*, Practical secure aggregation for privacy-preserving machine learning, presented at the Proceedings of the 2017 ACM SIGSAC Conference on Computer and Communications Security, 2017, 2017.

[13] Y. Liu, Y. Kang, C. Xing, T. Chen, and Q. Yang, A secure federated transfer learning framework, *IEEE Intelligent Systems,* **35** (4) (2020), pp. 70-82.

[14] V. Smith, C.-K. Chiang, M. Sanjabi, and A. S. Talwalkar, Federated multi-task learning, *Advances in neural information processing systems,* vol. 30, 2017.

[15] Y. Zhao, M. Li, L. Lai, N. Suda, D. Civin, and V. Chandra, Federated learning with non-iid data, *arXiv preprint arXiv:1806.00582,* 2018.

[16] R. C. Geyer, T. Klein, and M. Nabi, Differentially private federated learning: A client level perspective, *arXiv preprint arXiv:1712.07557,* 2017.

[17] M. Luo *et al.*, Metaselector: Meta-learning for recommendation with user-level adaptive model selection, presented at the Proceedings of The Web Conference 2020, 2020.

[18] L. Li, Y. Fan, M. Tse, and K.-Y. Lin, A review of applications in federated learning, *Computers & Industrial Engineering,* vol. 149, pp. 106854, 2020.

[19] P. Mohassel and Y. Zhang, Secureml: A system for scalable privacy-preserving machine learning, in *2017 IEEE symposium on security and privacy (SP),* 2017, pp. 19-38: IEEE.

[20] N. Kilbertus, A. Gascón, M. Kusner, M. Veale, K. Gummadi, and A. Weller, Blind justice: Fairness with encrypted sensitive attributes, in *International Conference on Machine Learning*, 2018, pp. 2630-2639: PMLR.

[21] D. Bogdanov, S. Laur, and J. Willemson, Sharemind: A framework for fast privacy-preserving computations, pp. 192-206: Springer, 2018.

[22] P. Mohassel and P. Rindal, ABY3: A mixed protocol framework for machine learning, in *Proceedings of the 2018 ACM SIGSAC conference on computer and communications security*, 2018, pp. 35-52.

[23] T. Araki, J. Furukawa, Y. Lindell, A. Nof, and K. Ohara, High-throughput semi-honest secure three-party computation with an honest majority, in *Proceedings of the 2016 ACM SIGSAC Conference on Computer and Communications Security*, 2016, pp. 805-817.

[24] P. Mohassel, M. Rosulek, and Y. Zhang, Fast and secure three-party computation: The garbled circuit approach, in *Proceedings of the 22nd ACM SIGSAC Conference on Computer and Communications Security*, 2015, pp. 591-602.

[25] R. L. Rivest, L. Adleman, and M. L. Dertouzos, On data banks and privacy homomorphisms, *Foundations of secure computation,* **4** (11) (1978), pp. 169-180.

[26] I. Giacomelli, S. Jha, M. Joye, C. D. Page, and K. Yoon, Privacy-preserving ridge regression with only linearly-homomorphic encryption, in *International Conference on Applied Cryptography and Network Security*, 2018, pp. 243-261: Springer.

[27] R. Hall, S. E. Fienberg, and Y. Nardi, Secure multiple linear regression based on homomorphic encryption, *Journal of Official Statistics,* 27 (4) (2011). pp. 669.

[28] V. Nikolaenko, U. Weinsberg, S. Ioannidis, M. Joye, D. Boneh, and N. Taft, Privacy-preserving ridge regression on hundreds of millions of records, in *2013 IEEE symposium on security and privacy*, 2013, pp. 334-348: IEEE.

[29] J. Yuan and S. Yu, Privacy preserving back-propagation neural network learning made practical with cloud computing, *IEEE Transactions on Parallel and Distributed Systems,* 25(1) (2013), 212-221.

[30] Q. Zhang, L. T. Yang, and Z. Chen, Privacy preserving deep computation model on cloud for big data feature learning, *IEEE Transactions on Computers,* **65**(5) (2015), 1351-1362.

[31] A. Acar, H. Aksu, A. S. Uluagac, and M. Conti, A survey on homomorphic encryption schemes: Theory and implementation, *ACM Computing Surveys (Csur),* 51 (4) (2018), 1-35.

[32] Y. Aono, T. Hayashi, L. Trieu Phong, and L. Wang, Scalable and secure logistic regression via homomorphic encryption, in *Proceedings of the Sixth ACM Conference on Data and Application Security and Privacy*, 2016, pp. 142-144.

[33] M. Kim, Y. Song, S. Wang, Y. Xia, and X. Jiang, Secure logistic regression based on homomorphic encryption: Design and evaluation, *JMIR medical informatics,* vol. 6, no. 2, p. e8805, 2018.

[34] C. Dwork, Differential privacy: A survey of results, in *International conference on theory and applications of models of computation*, 2008, pp. 1-19: Springer.

[35] Y. Aono, T. Hayashi, L. Wang, and S. Moriai, Privacy-preserving deep learning via additively homomorphic encryption, *IEEE Transactions on Information Forensics and Security,* vol. 13, no. 5, pp. 1333-1345, 2017.

[36] E. Bagdasaryan, A. Veit, Y. Hua, D. Estrin, and V. Shmatikov, How to backdoor federated learning, in *International Conference on Artificial Intelligence and Statistics*, 2020, pp. 2938-2948: PMLR.

[37] L. Melis, C. Song, E. De Cristofaro, and V. Shmatikov, Inference attacks against collaborative learning, *arXiv preprint arXiv:1805.04049,* vol. 13, 2018.

[38] L. Su and J. Xu, Securing distributed machine learning in high dimensions, *arXiv preprint arXiv:1804.10140,* pp. 1536-1233, 2018.

[39] H. Kim, J. Park, M. Bennis, and S.-L. Kim, On-device federated learning via blockchain and its latency analysis, *arXiv preprint arXiv:1808.03949,* 2018.

[40] S. J. Pan and Q. Yang, A survey on transfer learning, *IEEE Transactions on knowledge and data engineering,* vol. 22, no. 10, pp. 1345-1359, 2009.

[41] L. Wan, W. K. Ng, S. Han, and V. C. S. Lee, Privacy-preservation for gradient descent methods, in *Proceedings of the 13th ACM SIGKDD international conference on Knowledge discovery and data mining*, 2007, pp. 775-783.

[42] W. Du, Y. S. Han, and S. Chen, Privacy-preserving multivariate statistical analysis: Linear regression and classification, in *Proceedings of the 2004 SIAM international conference on data mining*, 2004, pp. 222-233: SIAM.

[43] A. Gascón *et al.*, Secure Linear Regression on Vertically Partitioned Datasets, *IACR Cryptol. ePrint Arch.,* vol. 2016, p. 892, 2016.

[44] A. F. Karr, X. Lin, A. P. Sanil, and J. P. Reiter, Privacy-preserving analysis of vertically partitioned data using secure matrix products, *Journal of Official Statistics,* vol. 25, no. 1, p. 125, 2009.

[45] A. P. Sanil, A. F. Karr, X. Lin, and J. P. Reiter, Privacy preserving regression modelling via distributed computation, in *Proceedings of the tenth ACM SIGKDD international conference on Knowledge discovery and data mining*, 2004, pp. 677-682.

[46] J. Vaidya and C. Clifton, Privacy preserving association rule mining in vertically partitioned data, in *Proceedings of the eighth ACM SIGKDD international conference on Knowledge discovery and data mining*, 2002, pp. 639-644.

[47] W. Du and M. J. Atallah, Privacy-preserving cooperative statistical analysis, in *Seventeenth Annual Computer Security Applications Conference*, 2001, pp. 102-110: IEEE.

[48] S. Hardy *et al.*, Private federated learning on vertically partitioned data via entity resolution and additively homomorphic encryption, *arXiv preprint arXiv:1711.10677,* 2017.

[49] R. Nock *et al.*, Entity resolution and federated learning get a federated resolution, *arXiv preprint arXiv:1803.04035,* 2018.

[50] O. Goldreich, S. Micali, and A. Wigderson, How to play any mental game, or a completeness theorem for protocols with honest majority, in *Providing Sound Foundations for Cryptography: On the Work of Shafi Goldwasser and Silvio Micali*, 2019, pp. 307-328.

[51] R. Shokri and V. Shmatikov, Privacy-preserving deep learning, in *Proceedings of the 22nd ACM SIGSAC conference on computer and communications security*, 2015, pp. 1310-1321.

[52] H. B. McMahan, E. Moore, D. Ramage, and B. A. y Arcas, Federated learning of deep networks using model averaging, *arXiv preprint arXiv:1602.05629,* 2016.

[53] J. Konečný, H. B. McMahan, D. Ramage, and P. Richtárik, Federated optimization: Distributed machine learning for on-device intelligence, *arXiv preprint arXiv:1610.02527,* 2016.

[54] Y. Lin, S. Han, H. Mao, Y. Wang, and W. J. Dally, Deep gradient compression: Reducing the communication bandwidth for distributed training, *arXiv preprint arXiv:1712.01887,* 2017.

[55] X. Ma, J. Ma, H. Li, Q. Jiang, and S. Gao, PDLM: Privacy-preserving deep learning model on cloud with multiple keys, *IEEE Transactions on Services Computing,* 2018.

[56] B. Hitaj, G. Ateniese, and F. Perez-Cruz, Deep models under the GAN: information leakage from collaborative deep learning, in *Proceedings of the 2017 ACM SIGSAC conference on computer and communications security*, 2017, pp. 603-618.

[57] Q. Yang, Y. Liu, T. Chen, and Y. Tong, Federated machine learning: Concept and applications, *ACM Transactions on Intelligent Systems and Technology (TIST),* vol. 10, no. 2, pp. 1-19, 2019.

[58] C. Ma *et al.*, On safeguarding privacy and security in the framework of federated learning, *IEEE network,* vol. 34, no. 4, pp. 242-248, 2020.

[59] S. Niknam, H. S. Dhillon, and J. H. Reed, Federated learning for wireless communications: Motivation, opportunities, and challenges, *IEEE Communications Magazine,* vol. 58, no. 6, pp. 46-51, 2020.

[60] M. M. Amiri and D. Gündüz, Federated learning over wireless fading channels, *IEEE Transactions on Wireless Communications,* vol. 19, no. 5, pp. 3546-3557, 2020.

[61] H.-V. Tran, G. Kaddoum, H. Elgala, C. Abou-Rjeily, and H. Kaushal, Lightwave power transfer for federated learning-based wireless networks, *IEEE Communications Letters,* vol. 24, no. 7, pp. 1472-1476, 2020.

[62] F. Ang, L. Chen, N. Zhao, Y. Chen, W. Wang, and F. R. Yu, Robust federated learning with noisy communication, *IEEE Transactions on Communications,* vol. 68, no. 6, pp. 3452-3464, 2020.

[63]   T. D. Nguyen, S. Marchal, M. Miettinen, M. H. Dang, N. Asokan, and A. R. Sadeghi, A crowdsourced self-learning approach for detecting compromised IoT devices, *arXiv preprint, arXiv: 1804.07474,* 2018.

[64]   S. Savazzi, M. Nicoli, and V. Rampa, Federated learning with cooperating devices: A consensus approach for massive IoT networks, *IEEE Internet of Things Journal,* vol. 7, no. 5, pp. 4641-4654, 2020.

[65]   M. S. H. Abad, E. Ozfatura, D. Gunduz, and O. Ercetin, Hierarchical federated learning across heterogeneous cellular networks, in *ICASSP 2020-2020 IEEE International Conference on Acoustics, Speech and Signal Processing (ICASSP)*, 2020, pp. 8866-8870: IEEE.

[66]   Y. Mansour, M. Mohri, J. Ro, and A. T. Suresh, Three approaches for personalization with applications to federated learning, *arXiv preprint arXiv:2002.10619,* 2020.

[67]   N. Rieke *et al.*, The future of digital health with federated learning, *NPJ digital medicine,* vol. 3, no. 1, pp. 1-7, 2020.

[68]   W. N. Price and I. G. Cohen, Privacy in the age of medical big data, *Nature medicine,* vol. 25, no. 1, pp. 37-43, 2019.

[69]   J. S. H. Lee *et al.*, From Discovery to Practice and Survivorship: Building a National Real‑World Data Learning Healthcare Framework for Military and Veteran Cancer Patients, *Clinical pharmacology and therapeutics,* vol. 106, no. 1, p. 52, 2019.

[70]   S. Bharati, P. Podder, M. R. H. Mondal, V. B. S. Prasath, and N. Gandhi, Ensemble Learning for Data-Driven Diagnosis of Polycystic Ovary Syndrome, presented at the 21st International Conference on Intelligent Systems Design and Applications (ISDA 2021) 2021.

[71]   W. Zhang *et al.*, Dynamic-Fusion-Based Federated Learning for COVID-19 Detection, *IEEE Internet of Things Journal,* vol. 8, no. 21, pp. 15884-15891, 2021.

[72]   R. Kumar *et al.*, Blockchain-federated-learning and deep learning models for covid-19 detection using ct imaging, *IEEE Sensors Journal,* vol. 21, no. 14, pp. 16301-16314, 2021.

[73]   P. Podder, S. Bharati, M. R. H. Mondal, and U. Kose, Application of Machine Learning for the Diagnosis of COVID-19, in *Data Science for COVID-19*: Elsevier, 2021, pp. 175-194.

[74]   M. R. H. Mondal, S. Bharati, and P. Podder, CO-IRv2: Optimized InceptionResNetV2 for COVID-19 detection from chest CT images, *PloS one,* vol. 16, no. 10, p. e0259179, 2021.

[75]   S. Bharati, P. Podder, M. Mondal, and V. B. Prasath, CO-ResNet: Optimized ResNet model for COVID-19 diagnosis from X-ray images, *International Journal of Hybrid Intelligent Systems,* vol. 17, pp. 71-85, 2021.

[76]   M. R. H. Mondal, S. Bharati, P. Podder, and P. Podder, Data analytics for novel coronavirus disease, *Informatics in Medicine Unlocked,* vol. 20, p. 100374, 2020.

[77]   M. R. H. Mondal, S. Bharati, and P. Podder, Diagnosis of COVID-19 Using Machine Learning and Deep Learning: A Review, *Current Medical Imaging,* 2021.

[78]   P. Podder, A. Khamparia, M. R. H. Mondal, M. A. Rahman, and S. Bharati, Forecasting the Spread of COVID-19 and ICU Requirements, *International Journal of Online and Biomedical Engineering (iJOE),* no. 5, 2021.

[79]   S. Bharati, P. Podder, and M. R. H. Mondal, Hybrid deep learning for detecting lung diseases from X-ray images, *Informatics in Medicine Unlocked,* p. 100391, 2020.

[80]   S. Bharati, P. Podder, M. R. H. Mondal, and V. B. Prasath, Medical Imaging with Deep Learning for COVID-19 Diagnosis: A Comprehensive Review, *International Journal of Computer Information Systems and Industrial Management Applications,* vol. 13, pp. 91 - 112, 2021.

[81] S. Bharati, P. Podder, M. R. H. Mondal, and N. Gandhi, Optimized NASNet for Diagnosis of COVID-19 from Lung CT Images, presented at the 20th International Conference on Intelligent Systems Design and Applications (ISDA 2020), 2020.

[82] S. Bharati, P. Podder, M. R. H. Mondal, P. Podder, and U. Kose, A review on epidemiology, genomic characteristics, spread, and treatments of COVID-19, in *Data Science for COVID-19*: Elsevier, 2022, pp. 487-505.

[83] P. K. Paul, S. Bharati, P. Podder, and M. R. Hossain Mondal, 10 The role of IoMT during pandemics Computational Intelligence for Managing Pandemics, A. Khamparia, R. Hossain Mondal, P. Podder, B. Bhushan, V. H. C. d. Albuquerque, and S. Kumar, Eds.: De Gruyter, 2021, pp. 169-186.

[84] S. Bharati and M. R. Hossain Mondal, 12 Applications and challenges of AI-driven IoHT for combating pandemics: a review Computational Intelligence for Managing Pandemics, A. Khamparia, R. Hossain Mondal, P. Podder, B. Bhushan, V. H. C. d. Albuquerque, and S. Kumar, Eds.: De Gruyter, 2021, pp. 213-230.

[85] K. V. Sarma *et al.*, Federated learning improves site performance in multicenter deep learning without data sharing, *Journal of the American Medical Informatics Association,* vol. 28, no. 6, pp. 1259-1264, 2021.

[86] S. Bharati and P. Podder, 1 Performance of CNN for predicting cancerous lung nodules using LightGBM, in *Artificial Intelligence for Data-Driven Medical Diagnosis*: De Gruyter, 2021, pp. 1-18.

[87] P. Podder, S. Bharati, and M. R. H. Mondal, 10 Automated gastric cancer detection and classification using machine learning, in *Artificial Intelligence for Data-Driven Medical Diagnosis*: De Gruyter, 2021, pp. 207-224.

[88] S. Bharati, P. Podder, and M. R. H. Mondal, Artificial Neural Network Based Breast Cancer Screening: A Comprehensive Review, *International Journal of Computer Information Systems and Industrial Management Applications,* vol. 12, pp. 125-137, 2020.

[89] S. Bharati, M. A. Rahman, and P. Podder, Breast cancer prediction applying different classification algorithm with comparative analysis using WEKA, in *2018 4th International Conference on Electrical Engineering and Information & Communication Technology (iCEEiCT)*, Dhaka, Bangladesh, 2018, pp. 581-584: IEEE.

[90] S. Bharati, P. Podder, R. Mondal, A. Mahmood, and M. Raihan-Al-Masud, Comparative performance analysis of different classification algorithm for the purpose of prediction of lung cancer, in *International Conference on Intelligent Systems Design and Applications*, 2018, vol. 941, pp. 447-457: Springer.

[91] A. Khamparia *et al.*, Diagnosis of Breast Cancer Based on Modern Mammography using Hybrid Transfer Learning, *Multidimensional Systems and Signal Processing,* 2020.

[92] S. Bharati, How Artificial Intelligence Impacts Businesses in the Period of Pandemics?, *Journal of the International Academy for Case Studies,* Editorials vol. 26, no. 5, 2020.

[93] S. Bharati, P. Podder, and P. K. Paul, Lung cancer recognition and prediction according to random forest ensemble and RUSBoost algorithm using LIDC data, *International Journal of Hybrid Intelligent Systems,* vol. 15, no. 2, pp. 91-100, 2019.

[94] P. Podder, S. Bharati, M. A. Rahman, and U. Kose, Transfer Learning for Classification of Brain Tumor, in *Deep Learning for Biomedical Applications*: CRC Press, 2021, pp. 315-328.

[95] M. Aledhari, R. Razzak, R. M. Parizi, and F. Saeed, Federated learning: A survey on enabling technologies, protocols, and applications, *IEEE Access,* vol. 8, pp. 140699-140725, 2020.

[96]  S. Bharati, P. Podder, and M. Raihan-Al-Masud, EEG Eye State Prediction and Classification in order to Investigate Human Cognitive State, In 2018 *International Conference on Advancement in Electrical and Electronic Engineering* (ICAEEE), pp. 1-4. IEEE, 2018.

[97]  M. Sadiq Iqbal, N. Akhtar, A. H. M. Shahariar Parvez, S. Bharati, and P. Podder, Ensemble learning-based EEG feature vector analysis for brain computer interface, in *Evolutionary Computing and Mobile Sustainable Networks*: Springer, 2021, pp. 957-969.

[98]  H. Elayan, M. Aloqaily, and M. Guizani, Sustainability of healthcare data analysis IoT-based systems using deep federated learning, *IEEE Internet of Things Journal,* 2021.

[99]  Z. Xiao, X. Xu, H. Xing, F. Song, X. Wang, and B. Zhao, A federated learning system with enhanced feature extraction for human activity recognition, *Knowledge-Based Systems,* vol. 229, p. 107338, 2021.

[100] P. Chhikara, P. Singh, R. Tekchandani, N. Kumar, and M. Guizani, Federated learning meets human emotions: A decentralized framework for human–computer interaction for IoT applications, *IEEE Internet of Things Journal,* vol. 8, no. 8, pp. 6949-6962, 2020.

[101] A. Vaid *et al.*, Federated learning of electronic health records to improve mortality prediction in hospitalized patients with COVID-19: Machine learning approach, *JMIR medical informatics,* vol. 9, no. 1, p. e24207, 2021.

[102] T. S. Brisimi, R. Chen, T. Mela, A. Olshevsky, I. C. Paschalidis, and W. Shi, Federated learning of predictive models from federated electronic health records, *International journal of medical informatics,* vol. 112, pp. 59-67, 2018.

[103] B. S. Glicksberg, K. W. Johnson, and J. T. Dudley, The next generation of precision medicine: observational studies, electronic health records, biobanks and continuous monitoring, *Human molecular genetics,* vol. 27, no. R1, pp. R56-R62, 2018.

[104] P. B. Jensen, L. J. Jensen, and S. Brunak, Mining electronic health records: towards better research applications and clinical care, *Nature Reviews Genetics,* vol. 13, no. 6, pp. 395-405, 2012.

[105] R. Miotto, F. Wang, S. Wang, X. Jiang, and J. T. Dudley, Deep learning for healthcare: review, opportunities and challenges, *Briefings in bioinformatics,* vol. 19, no. 6, pp. 1236-1246, 2018.

[106] Z. Obermeyer, B. Powers, C. Vogeli, and S. Mullainathan, Dissecting racial bias in an algorithm used to manage the health of populations, *Science,* vol. 366, no. 6464, pp. 447-453, 2019.

[107] G. Hripcsak *et al.*, Observational Health Data Sciences and Informatics (OHDSI): opportunities for observational researchers, *Studies in health technology and informatics,* vol. 216, p. 574, 2015.

[108] D. Yang *et al.*, Federated semi-supervised learning for COVID region segmentation in chest CT using multi-national data from China, Italy, Japan, *Medical image analysis,* vol. 70, p. 101992, 2021.

[109] R. Duan *et al.*, Learning from electronic health records across multiple sites: A communication-efficient and privacy-preserving distributed algorithm, *Journal of the American Medical Informatics Association,* vol. 27, no. 3, pp. 376-385, 2020.

[110] A. M. Begum, M. R. H. Mondal, P. Podder, and S. Bharati, Detecting Spinal Abnormalities using Multilayer Perceptron Algorithm, presented at the 11th World Congress on Information and Communication Technologies, 2021.

[111] L. Huang, A. L. Shea, H. Qian, A. Masurkar, H. Deng, and D. Liu, Patient clustering improves efficiency of federated machine learning to predict mortality and hospital stay time using distributed electronic medical records, *Journal of biomedical informatics,* vol. 99, p. 103291, 2019.

[112] Z. Li, K. Roberts, X. Jiang, and Q. Long, Distributed learning from multiple EHR databases: contextual embedding models for medical events, *Journal of biomedical informatics,* vol. 92, p. 103138, 2019.

[113] P. V. Raja and E. Sivasankar, Modern framework for distributed healthcare data analytics based on Hadoop, in *Information and Communication Technology-EurAsia Conference*, 2014, pp. 348-355: Springer.

[114] Y. Kim, J. Sun, H. Yu, and X. Jiang, Federated tensor factorization for computational phenotyping, in *Proceedings of the 23rd ACM SIGKDD International Conference on Knowledge Discovery and Data Mining*, 2017, pp. 887-895.

[115] J. Lee, J. Sun, F. Wang, S. Wang, C.-H. Jun, and X. Jiang, Privacy-preserving patient similarity learning in a federated environment: development and analysis, *JMIR medical informatics,* vol. 6, no. 2, p. e7744, 2018.

[116] D. Liu, D. Dligach, and T. Miller, Two-stage federated phenotyping and patient representation learning, in *Proceedings of the conference. Association for Computational Linguistics. Meeting*, 2019, vol. 2019, p. 283: NIH Public Access.

[117] O. Gupta and R. Raskar, Distributed learning of deep neural network over multiple agents, *Journal of Network and Computer Applications,* vol. 116, pp. 1-8, 2018.

[118] P. Vepakomma, O. Gupta, T. Swedish, and R. Raskar, Split learning for health: Distributed deep learning without sharing raw patient data, *arXiv preprint arXiv:1812.00564,* 2018.

[119] S. Silva, B. A. Gutman, E. Romero, P. M. Thompson, A. Altmann, and M. Lorenzi, Federated learning in distributed medical databases: Meta-analysis of large-scale subcortical brain data, pp. 270-274: IEEE.

[120] S. Boughorbel, F. Jarray, N. Venugopal, S. Moosa, H. Elhadi, and M. Makhlouf, Federated uncertainty-aware learning for distributed hospital ehr data, *arXiv preprint arXiv:1910.12191,* 2019.

[121] P. Sharma, F. E. Shamout, and D. A. Clifton, Preserving patient privacy while training a predictive model of in-hospital mortality, *arXiv preprint arXiv:1912.00354,* 2019.

[122] S. AbdulRahman, H. Tout, H. Ould-Slimane, A. Mourad, C. Talhi, and M. Guizani, A survey on federated learning: The journey from centralized to distributed on-site learning and beyond, *IEEE Internet of Things Journal,* vol. 8, no. 7, pp. 5476-5497, 2020.

[123] S. R. Pfohl, A. M. Dai, and K. Heller, Federated and differentially private learning for electronic health records, *arXiv preprint arXiv:1911.05861,* 2019.

[124] Y. Chen, X. Qin, J. Wang, C. Yu, and W. Gao, Fedhealth: A federated transfer learning framework for wearable healthcare, *IEEE Intelligent Systems,* vol. 35, no. 4, pp. 83-93, 2020.

[125] B. Yuan, S. Ge, and W. Xing, A federated learning framework for healthcare iot devices, *arXiv preprint arXiv:2005.05083,* 2020.

[126] A. Vaid *et al.*, Federated learning of electronic health records improves mortality prediction in patients hospitalized with covid-19, *medRxiv,* 2020.

[127] H. B. McMahan, D. Ramage, K. Talwar, and L. Zhang, Learning differentially private recurrent language models, *arXiv preprint arXiv:1710.06963,* 2017.

[128] C. H. Van Berkel, Multi-core for mobile phones, in *2009 Design, Automation & Test in Europe Conference & Exhibition*, 2009, pp. 1260-1265: IEEE.

[129] J. Kang, Z. Xiong, D. Niyato, Y. Zou, Y. Zhang, and M. Guizani, Reliable federated learning for mobile networks, *IEEE Wireless Communications,* vol. 27, no. 2, pp. 72-80, 2020.

[130] J. Li, M. Khodak, S. Caldas, and A. Talwalkar, Differentially private meta-learning, *arXiv preprint arXiv:1909.05830,* 2019.

[131] T. Li, M. Sanjabi, A. Beirami, and V. Smith, Fair resource allocation in federated learning, *arXiv preprint arXiv:1905.10497,* 2019.
[132] Y. Laguel, K. Pillutla, J. Malick, and Z. Harchaoui, Device heterogeneity in federated learning: A superquantile approach, *arXiv preprint arXiv:2002.11223,* 2020.
[133] C. T Dinh, N. Tran, and J. Nguyen, Personalized federated learning with moreau envelopes, *Advances in Neural Information Processing Systems,* vol. 33, pp. 21394-21405, 2020.
[134] M. Mohri, G. Sivek, and A. T. Suresh, Agnostic federated learning, in *International Conference on Machine Learning*, 2019, pp. 4615-4625: PMLR.
[135] P. Awasthi, C. Cortes, Y. Mansour, and M. Mohri, Beyond individual and group fairness, *arXiv preprint arXiv:2008.09490,* 2020.
[136] M. Hardt, E. Price, and N. Srebro, Equality of opportunity in supervised learning, *Advances in neural information processing systems,* vol. 29, 2016.
[137] T. Hashimoto, M. Srivastava, H. Namkoong, and P. Liang, Fairness without demographics in repeated loss minimization, in *International Conference on Machine Learning*, 2018, pp. 1929-1938: PMLR.
[138] M. B. Zafar, I. Valera, M. G. Rogriguez, and K. P. Gummadi, Fairness constraints: Mechanisms for fair classification, in *Artificial Intelligence and Statistics*, 2017, pp. 962-970: PMLR.
[139] Y. Yao, L. Rosasco, and A. Caponnetto, On early stopping in gradient descent learning, *Constructive Approximation,* vol. 26, no. 2, pp. 289-315, 2007.
[140] L. Mackey, M. Jordan, and A. Talwalkar, Divide-and-conquer matrix factorization, *Advances in neural information processing systems,* vol. 24, 2011.
[141] T. Bolukbasi, J. Wang, O. Dekel, and V. Saligrama, Adaptive neural networks for efficient inference, in *International Conference on Machine Learning*, 2017, pp. 527-536: PMLR.
[142] D. Yin, A. Pananjady, M. Lam, D. Papailiopoulos, K. Ramchandran, and P. Bartlett, Gradient diversity: a key ingredient for scalable distributed learning, in *International Conference on Artificial Intelligence and Statistics*, 2018, pp. 1998-2007: PMLR.
[143] N. Agarwal, A. T. Suresh, F. X. X. Yu, S. Kumar, and B. McMahan, cpSGD: Communication-efficient and differentially-private distributed SGD, *Advances in Neural Information Processing Systems,* vol. 31, 2018.
[144] H. Eichner, T. Koren, B. McMahan, N. Srebro, and K. Talwar, Semi-cyclic stochastic gradient descent, in *International Conference on Machine Learning*, 2019, pp. 1764-1773: PMLR.
[145] K. Bonawitz, H. Eichner, and W. Grieskamp, TensorFlow federated: machine learning on decentralized data, (2020).
[146] S. Caldas *et al.*, Leaf: A benchmark for federated settings, *arXiv preprint arXiv:1812.01097,* 2018.
[147] S. Bharati, P. Podder, M.R.H. Mondal, M.R.A. Robel, Threats and countermeasures of cyber security in direct and remote vehicle communication systems. Journal of Information Assurance and Security, **15** (4), pp. 153-164, 2020.
[148] S. Bharati, P. Podder, D. N. H. Thanh, V. B. S. Prasath. Dementia classification using MR imaging and voting based machine learning models, Multimedia Tools and Applications, 2022. https://doi.org/10.1007/s11042-022-12754-x
[149] S. Bharati, P. Podder, Adaptive PAPR reduction scheme for OFDM using SLM with the fusion of proposed clipping and filtering technique in order to diminish PAPR and signal distortion. Wireless Personal Communications, 113, no. 4 (2020): 2271-2288.
[150] P. Podder, S. Bharati, M.R.H. Mondal, P. K. Paul, and U. Kose. Artificial neural network for cybersecurity: a comprehensive review. Journal of Information Assurance and Security, **16** (1), pp. 010 – 023, 2021.

[151]    S. Bharati, P. Podder, and M. R. H. Mondal. Diagnosis of polycystic ovary syndrome using machine learning algorithms. In 2020 IEEE Region 10 Symposium (TENSYMP), pp. 1486-1489. IEEE, 2020.